\pdfoutput=1

\documentclass[11pt]{article}

\usepackage[final]{EMNLP2023}

\usepackage{times}
\usepackage{latexsym}
\usepackage{amssymb} 

\usepackage[T1]{fontenc}

\usepackage[utf8]{inputenc}

\usepackage{microtype}

\usepackage{inconsolata}
\usepackage{graphicx}

\newcommand{\multiline}[1]{\begin{tabular}{@{}c@{}}#1\end{tabular}}

\title{R-BI: Regularized Batched Inputs enhance Incremental Decoding Framework for Low-Latency Simultaneous Speech Translation}

\author{\multiline{Jiaxin Guo, Zhanglin Wu, Zongyao Li, Hengchao Shang, Daimeng Wei,  \\ Xiaoyu Chen, Zhiqiang Rao, Shaojun Li, Hao Yang} \\
  Huawei Translation Services Center, Beijing, China \\
  \tt \{guojiaxin1,wuzhanglin2,lizongyao,shanghengchao,weidaimeng,\\ 
  \tt chenxiaoyu35,raozhiqiang,lishaojun18,yanghao30\}@huawei.com
  }

\begin{document}
\maketitle
\begin{abstract}

Incremental Decoding is an effective framework that enables the use of an offline model in a simultaneous setting without modifying the original model, making it suitable for Low-Latency Simultaneous Speech Translation. However, this framework may introduce errors when the system outputs from incomplete input. To reduce these output errors, several strategies such as Hold-$n$, LA-$n$, and SP-$n$ can be employed, but the hyper-parameter $n$ needs to be carefully selected for optimal performance. Moreover, these strategies are more suitable for end-to-end systems than cascade systems. In our paper, we propose a new adaptable and efficient policy named "Regularized Batched Inputs". Our method stands out by enhancing input diversity to mitigate output errors. We suggest particular regularization techniques for both end-to-end and cascade systems. We conducted experiments on IWSLT Simultaneous Speech Translation (SimulST) tasks, which demonstrate that our approach achieves low latency while maintaining no more than 2 BLEU points loss compared to offline systems. Furthermore, our SimulST systems attained several new state-of-the-art results in various language directions.

\end{abstract}

\section{Introduction}

\begin{figure}[!th]
\centering
\includegraphics[width=0.45\textwidth]{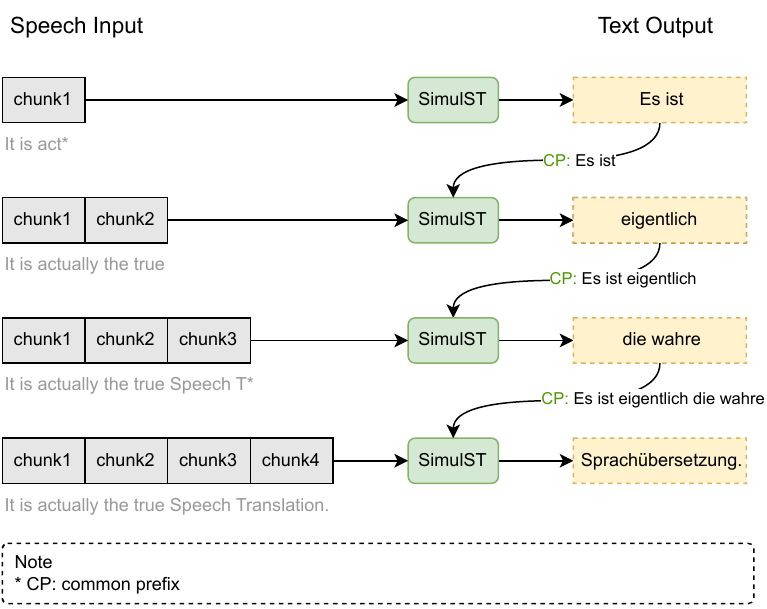} 
\caption{Incremental Decoding Framework}
\label{fig:ic}
\end{figure}

Simultaneous Speech Translation (SimulST, also known as real-time or streaming speech translation) \citep{DBLP:journals/mt/FugenWK07,DBLP:conf/acl/OdaNSTN14,DBLP:conf/naacl/DalviDSV18,DBLP:conf/emnlp/LiuDLLC21,DBLP:conf/acl/ZhangF22} is the task of generating text translations incrementally given partial speech input only. The goal of SimulST is to provide incremental translations as speech is being received, while minimizing the amount of latency. This trade-off between accuracy and latency is particularly critical in SimulST, where even a small delay could significantly impact the quality of the translation.

Conventional approaches to SimulST involve a cascade system that divides the translation process into two stages: simultaneous automatic speech recognition (ASR) \citep{DBLP:conf/asru/RaoSP17,DBLP:journals/corr/abs-1211-3711} and simultaneous neural machine translation (SimulMT) \citep{DBLP:conf/eacl/NeubigCGL17,DBLP:conf/icassp/LawsonCTRSJ18,DBLP:conf/iclr/MaPCPG20,DBLP:conf/acl/ArivazhaganCMCY19,DBLP:conf/iwslt/ArivazhaganCMF20,DBLP:conf/emnlp/LiuDLLC21,DBLP:conf/interspeech/ZaidiL0K22}. However, this approach suffers from increased latency and cannot be jointly optimized. Recent research efforts have focused on exploring end-to-end SimulST, as exemplified by the work of  \citet{DBLP:conf/ijcnlp/MaPK20} and \citet{DBLP:conf/acl/RenLTZQZL20}. These studies adapt SimulMT methods to SimulST. Specifically, \citet{DBLP:conf/ijcnlp/MaPK20}'s investigation delves into the impact of speech block processing on S2T simultaneous translation, while \citet{DBLP:conf/acl/RenLTZQZL20}'s experiments utilize a source language CTC segmenter. But, it is important to note that bilingual speech translation datasets, which are necessary for end-to-end models, are still scarce resources.


Recently, researchers focus on exploring Simultaneous Policies that directly apply Offline Speech Translation (OfflineST) \citep{DBLP:conf/icassp/LamSR21,DBLP:conf/acl/FangYLFW22} models to simultaneous scenarios. The overall framework of Simultaneous Policies is called Incremental Decoding \citep{DBLP:conf/naacl/DalviDSV18,DBLP:conf/interspeech/LiuSN20}, as shown in Figure \ref{fig:ic}. Notably, the recent IWSLT2022 competition \citep{DBLP:conf/iwslt/AnastasopoulosB22} was won using the one kind of Simultaneous Policy methods. The main advantage of this approach is that it eliminates the need for additional model training for simultaneous scenarios. \textbf{The key point of these policy designs is to bridge the gap between OfflineST and SimulST, where the biggest gap lies in the possible risk of introducing errors when the system outputs from incomplete input.} Specifically, two categories of policies exist: fixed and adaptive. While fixed policies, such as Hold-$n$ \citep{DBLP:conf/interspeech/LiuSN20,DBLP:conf/iwslt/PolakPNLMNBW22}, rely on simple heuristics, adaptive policies take input content into consideration when making decisions. Recent studies have demonstrated that adaptive policies, like LA-$n$ \citep{DBLP:conf/interspeech/LiuSN20,DBLP:conf/iwslt/PolakPNLMNBW22} and SP-$n$ \cite{DBLP:conf/interspeech/NguyenSW21,DBLP:conf/iwslt/PolakPNLMNBW22}, exhibit superior overall performance relative to their fixed counterparts. Moreover, with just a simple adaptation, the Simultaneous Policy can be used in both End-to-End and Cascaded systems. However, there are two main issues:
\begin{itemize}
    \item First, adaptive policies exhibit a direct correlation with the quantity of consecutive audio chunks, denoted as $n$. In essence, superior quality is achieved with increasing values of $n$; however, this leads to elevated latency \citep{DBLP:conf/iwslt/PolakPNLMNBW22}. Accordingly, in situations where low-latency is imperative, it becomes necessary to restrict $n$ to the range of $2-3$, which regrettably results in a deterioration of audio quality. To overcome this limitation, it would be advantageous to devise a more universal approach that is not reliant on the value of $n$.
    \item Secondly, in OfflineST \citep{DBLP:conf/icassp/LamSR21,DBLP:conf/acl/FangYLFW22}, cascade systems generally exhibit superior quality compared to end-to-end systems \citep{DBLP:conf/iwslt/AnsariABBCDDFFG20,DBLP:conf/iwslt/WangGQWWSCZTYQ22,DBLP:conf/iwslt/AnastasopoulosB22}. However, when using simulative policies to construct SimulST, we observe that end-to-end systems outperform cascade systems. This suggests that these policies are more amenable to end-to-end models than to cascade models. Is there a universal approach that can be employed for both cascade and end-to-end systems?
\end{itemize}

\textbf{The prior method selected the output from candidate results among consecutive inputs to minimize errors.} However, in our paper, we introduce a novel approach called Regularized Batched Inputs (R-BI). \textbf{R-BI focuses exclusively on the current input and generates a diverse array of inputs through regularization to enhance input diversity and reduce the likelihood of errors.} Our main contributions are as follows:

\begin{itemize}
    \item \textbf{Our proposed policy is both flexible and effective, capable of transforming OfflineST into SimulST while remaining robust to both end-to-end and cascaded systems.}
    \item \textbf{We have developed a detailed regularization method and validated its effectiveness.} For end-to-end systems, we propose multiple regularization methods specifically designed for speech-modality input. Similarly, for cascaded systems, we creatively have transformed the problem of regularizing text-modality input into diversifying the ASR outputs.
    \item To test the effectiveness of our R-BI policy, we trained two strong baseline end-to-end and cascaded OfflineST systems, which were then transformed into SimulST systems. \textbf{Moreover, our SimulST achieved a new state-of-the-art result.}
\end{itemize}

\section{Background}

\begin{figure*}[!th]
\centering
\includegraphics[width=0.9 \textwidth]{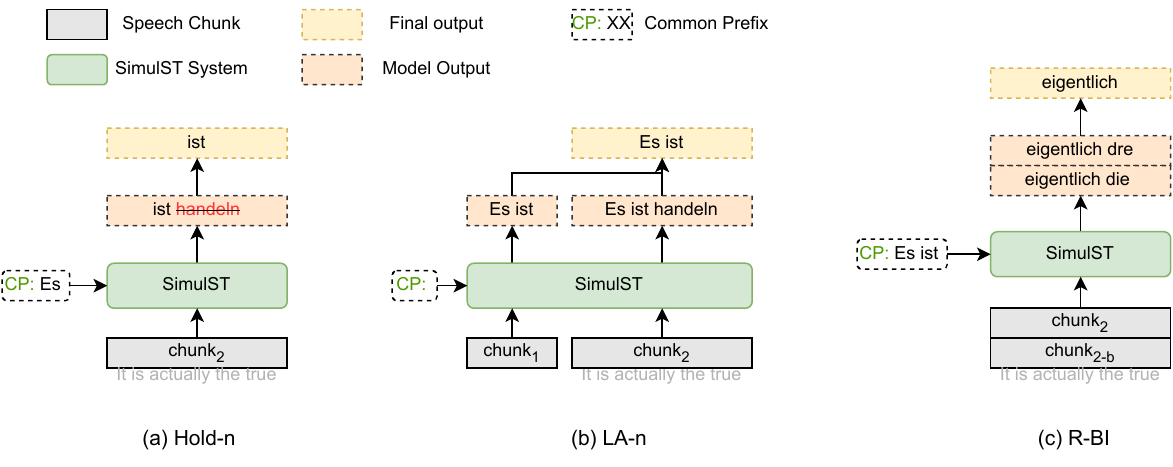} 
\caption{The difference between Hold-$n$ \citep{DBLP:conf/interspeech/LiuSN20}, LA-$n$ \citep{DBLP:conf/interspeech/LiuSN20} and our proposed R-BI. For example, (a) Hold-1 (b) LA-2 (c) R-BI with batch size = 2}
\label{fig:policy}
\end{figure*}

\subsection{Incremental Decoding}
Figure \ref{fig:ic} illustrates the Incremental Decoding framework \citep{DBLP:conf/naacl/DalviDSV18,DBLP:conf/interspeech/LiuSN20}. Translation tasks often require additional information that is not immediately apparent until the end of the input. While offline processing of the entire input at once typically yields superior results, this approach can introduce significant delays when working online. One potential solution to mitigate such latency is to partition the input into smaller segments and translate each one incrementally. This method reduces processing time while preserving the quality of the translation.

To achieve incremental inference, we divide the input into fixed-size chunks \citep{DBLP:conf/interspeech/NguyenSW21} and decode each segment as it arrives. Once a chunk has been selected, its predictions are finalized to avoid distractions from constantly changing hypotheses. The decoding of the next chunk depends on the committed predictions. In practice, decoding for new chunks can proceed from a previously buffered decoder state or begin after forced decoding with the tokens that have already been committed. Regardless, the source-target attention spans all available chunks rather than only the current one.

\subsection{Common Prefix Decisions Policy}
Making assumptions based on incomplete information can increase the likelihood of errors. To minimize this risk and prioritize accuracy over efficiency, it is recommended to implement the \textbf{Common Prefix Decisions Policy} \citep{DBLP:conf/iwslt/PolakPNLMNBW22}. There are two types of policy: one is fixed policy, such as Hold-$n$ \citep{DBLP:conf/interspeech/LiuSN20}; the other is more adaptive, such as LA-$n$ \citep{DBLP:conf/interspeech/LiuSN20}, SP-$n$ \citep{DBLP:conf/interspeech/NguyenSW21}. Here, we only discuss and compare the two most widely used strategies, Hold-$n$ and LA-$n$.


\paragraph{Hold-$n$} One of the simplest ways to select partial hypotheses is by either withholding or deleting the last $n$ tokens in each chunk. If $n$ exceeds the number of output tokens from the current chunk, the common prefix will be empty, resulting in no display for this chunk. Conversely, Hold-0 will fully output all chunk-level predictions. See Figure \ref{fig:policy}(a).



\paragraph{LA-$n$} Local agreement involves displaying the agreeing prefixes of $n$ consecutive chunks. During the first $n$-1 chunks, no tokens are output. Starting from the $n$-th chunk, we identify the longest common prefix of the best hypothesis from the $n$ consecutive chunks. See Figure \ref{fig:policy}(b).



\section{Approach: Regularized Batched Inputs}

\begin{figure*}[!th]
\centering
\includegraphics[width=\textwidth]{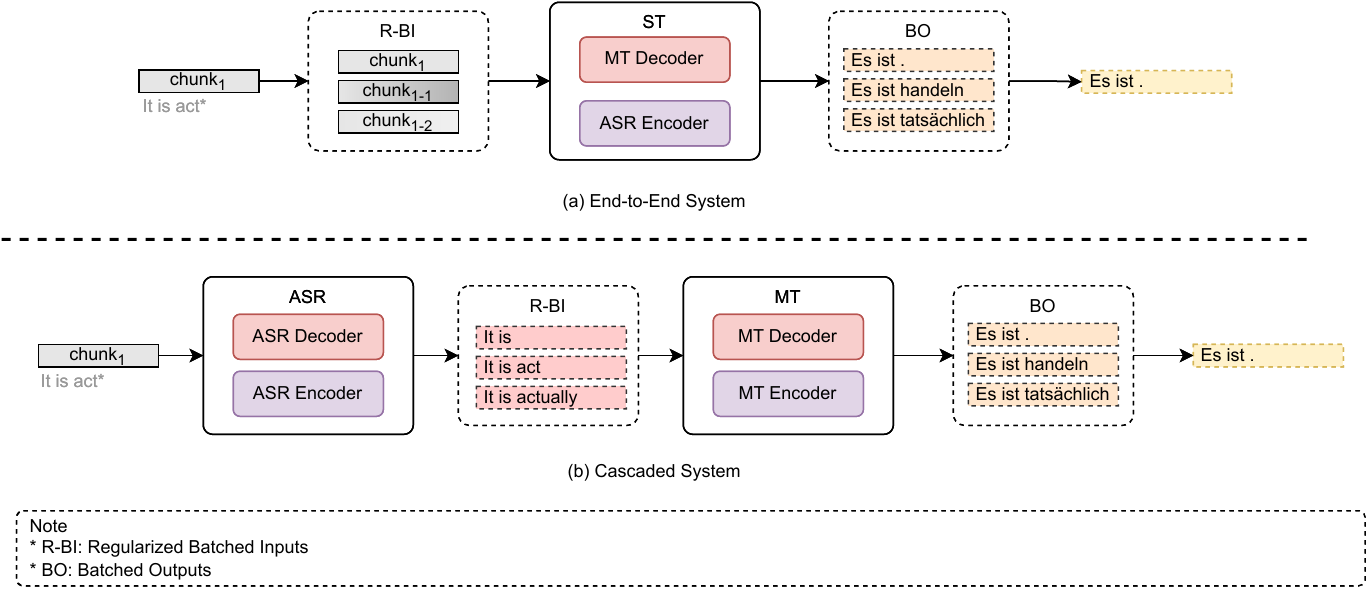} 
\caption{End-to-End and Cascaded Systems of R-BI.}
\label{fig:system}
\end{figure*}

The overall framework for our Regularized Batched Inputs (R-BI) method is illustrated in Figure \ref{fig:system}. The distinctions between our proposed approach and previous work are depicted in Figure \ref{fig:policy}. In the following sections, we will elaborate on the details of our method.

\subsection{Regularized Batched Inputs through Incremental Decoding}

At the decoding step $i$, given the incomplete input data $x_i$,  we apply multiple regularization methods $\{\mathcal{R}_1, \mathcal{R}_2, \ldots, \mathcal{R}_{B}\}$ to produce a set of regularized inputs. $B$ represent the number of regularization methods. These inputs are then combined with the original data, yielding a batch of regularized inputs, 
\begin{equation}
\textbf{x}_{[i:i+B+1]} = [x_i, x_{i+1}, \ldots, x_{i+B}]
\end{equation}
where $x_{i+b} = \mathcal{R}_{b}(x_i); b\in[1, B]$. We feed this batch into the model $\mathcal{P}(y|x)$, which produces a corresponding batch of outputs, 
\begin{equation}
\textbf{y}^\prime_{[i:i+B+1]} = [y^\prime_i, y^\prime_{i+1}, \ldots, y^\prime_{i+B}]
\end{equation}
where $y^\prime = \mathcal{P}(x)$. Within this batch of outputs, we identify the stable prefix $y_i$ that can guide the next incremental decoding step.
\begin{equation}
y = LCP(\textbf{y}^\prime_{[i:i+B+1]})
\end{equation}
where $LCP$ is longest common prefix of the arguments.



\begin{figure}[!th]
\centering
\includegraphics[width=0.45\textwidth]{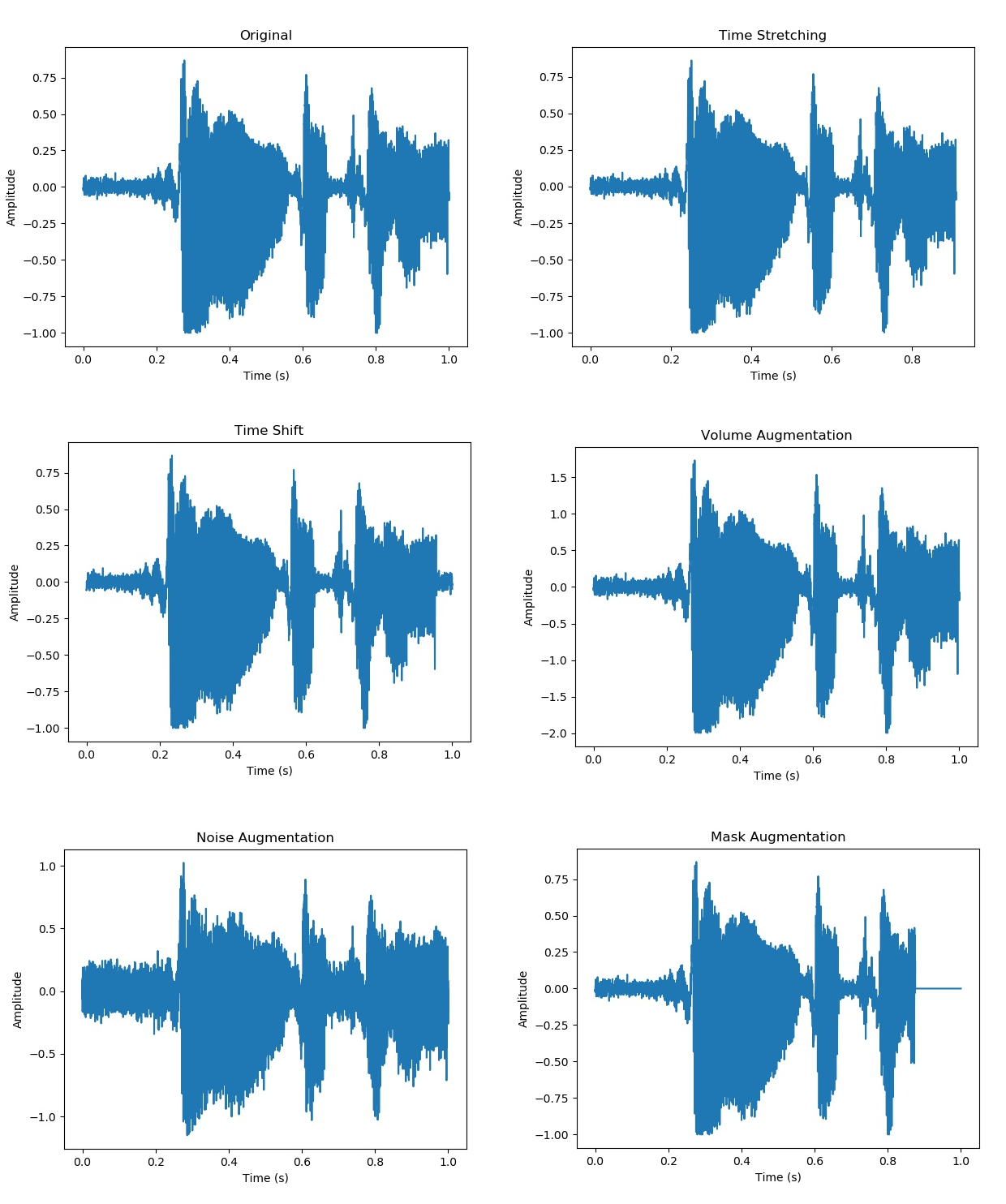} 
\caption{Different waveforms under different Regularization Methods for End-to-End System}
\label{fig:r_e2e}
\end{figure}

\subsection{End-to-End Regularization Method}

In End-to-End system, the input $x_i$ represents the speech information conveyed by the $i$-th set of audio chunks. Furthermore, we apply the regularization methods $\mathcal{R}$ to the speech input, as shown in Figure \ref{fig:system}(a). Specifically, we propose five methods as following:

\paragraph{Time Stretching} Time Stretching (TSt) is a technique used to change the speed of audio playback without affecting its pitch. This is achieved by applying linear interpolation to the audio samples, which involves resampling the audio data at a different rate to create a longer or shorter version of the original sound. The speed adjustment is typically controlled by specifying a range of minimum and maximum speeds, within which the actual speed is randomly selected.

\paragraph{Time Shift} Time Shift (TSh) \citep{DBLP:conf/interspeech/BasuIM98} is a technique used to randomly shift the audio signal in time, typically within a range of ±5\% of the signal length. This can be achieved by rolling the audio samples along the time axis using circular shifting in the spectral-domain. The amount of shift is randomly generated within the specified range and applied to the audio samples. The technique preserves all information in the signal and can be used for both one-dimensional audio data and multi-dimensional feature data with a sequence length and feature dimensionality.

\paragraph{Volume Augmentation} Volume Augmentation (VA) is a data augmentation technique that applies a random gain to the audio samples within a specified range. The gain values are exponentiated to create an exponential distribution.

\paragraph{Noise Augmentation} Noise Augmentation (NA) \citep{DBLP:conf/bmvc/PachoudGC08} is a technique used to introduce random noise into the audio signal. The technique typically involves applying different types of noise to the audio samples, such as uniform white noise or gaussian noise, with varying levels of intensity specified by the user. This can be achieved by generating random noise samples of a particular distribution and padding them to the original audio samples.

\paragraph{Mask Augmentation} Mask Augmentation (MA) is a technique that masks a portion of the input data. The implementation above applies time masking by setting a certain number of time steps to zero in the input.

\subsection{Cascaded Regularization Method}

In a cascaded system, an intriguing scenario arises where two inputs, represented by $x$ mentioned before, must be considered: the upstream raw speech input and the downstream direct text input. Unlike end-to-end systems, normalization of the speech input is not necessary; rather, direct normalization of the input text suffices. The crux of the matter, however, lies in the dual requirement to normalize the text while preserving any representation of the original speech information. To our knowledge, no comparable solutions have been proposed previously. After careful analysis, we discovered that the downstream MT input text is a byproduct of the upstream ASR output. Thus, this challenge can be addressed by generating multiple candidate text results via the upstream ASR. Furthermore, the candidate results produced by the ASR are inherently normalized, as shown in Figure \ref{fig:system}(b).


\begin{figure}[!th]
\centering
\includegraphics[width=0.45\textwidth]{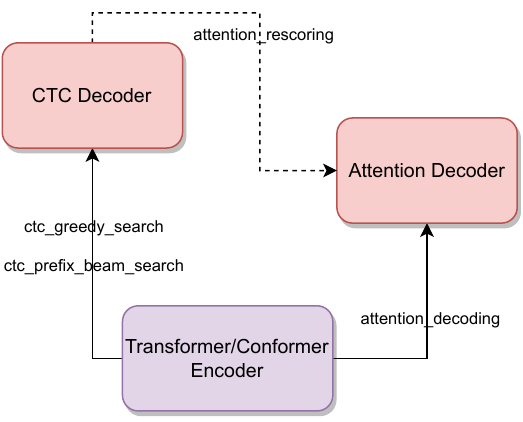} 
\caption{U2 Model: hybrid ASR structure combines a CTC decoder with an AED decoder}
\label{fig:u2}
\end{figure}

ASR systems have the ability to generate multiple hypotheses for each input speech utterance and use different decoding methods depending on their chosen decoder structure. To improve accuracy and efficiency, a hybrid ASR structure U2 \citep{DBLP:journals/corr/abs-2012-05481,DBLP:conf/icassp/KimHW17} seen in Figure \ref{fig:u2}, that combines the Connectionist Temporal Classification (CTC) decoder with an attention-based decoder(AED) has become widely used. The ASR Encoder can be Transformer \citep{DBLP:conf/nips/VaswaniSPUJGKP17} or Conformer \citep{DBLP:conf/interspeech/GulatiQCPZYHWZW20}. This hybrid structure offers several decoding methods including "attention\_decoding", "ctc\_greedy\_search", "ctc\_prefix\_beam\_search", and "attention\_rescoring". "attention\_decoding" mode is computationally expensive and cannot meet the time requirement of our current scenario. "ctc\_greedy\_search" only generates one result, which can not meet the input requirement of our current scenario. \textbf{In the cascaded system, we prefer "ctc\_prefix\_beam\_search" and "attention\_rescoring" decoding modes.}

\begin{itemize}
    \item \textbf{attention\_decoding}: apply standard autoregressive beam search on the AED part of the model.
    \item \textbf{ctc\_greedy\_search}: apply CTC greedy search on the CTC part of the model and output only one candidate.
    \item \textbf{ctc\_prefix\_beam\_search}: apply CTC prefix beam search on the CTC part of the model, which can give the n-best candidates.
    \item \textbf{attention\_rescoring}: first apply CTC prefix beam search on the CTC part of the model to generate n-best candidates, and then rescore the n-best candidates on the AED decoder part with corresponding encoder output.
\end{itemize}

\section{Experiments}

\subsection{Datasets}
In our experiments, we utilized the MuST-C V2 \citep{DBLP:conf/naacl/GangiCBNT19} dataset for English→German (EN→DE), English→Japanese (EN→JA) and English→Chinese (EN→ZH) speech translation. This multilingual dataset was recorded from TED talks and used solely for the English data in the ASR task. We validated using the dev set and reported our results on the tst-COMMON set. Additionally, we included three ASR datasets: LibriSpeech V12 \citep{DBLP:conf/icassp/PanayotovCPK15}, TEDLIUM V3 \citep{DBLP:conf/specom/HernandezNGTE18}, and CoVoST V2 \citep{DBLP:conf/interspeech/WangWGP21}. LibriSpeech \citep{DBLP:conf/icassp/PanayotovCPK15} is composed of audio book recordings with case-insensitive text lacking punctuation. In contrast, TEDLIUM \citep{DBLP:conf/specom/HernandezNGTE18} is a large-scale speech recognition dataset containing TED talk audio recordings along with text transcriptions. CoVoST \citep{DBLP:conf/interspeech/WangWGP21} is another multilingual speech translation dataset based on Common Voice, with opEN→domain content. Unlike LibriSpeech \citep{DBLP:conf/icassp/PanayotovCPK15}, both MuST-C \citep{DBLP:conf/naacl/GangiCBNT19} and CoVoST \citep{DBLP:conf/interspeech/WangWGP21} have case-sensitive text and punctuation.


\begin{table*}[!th]
    \centering
    {
    \begin{tabular}{lcccccc}
    \hline
    \hline
    Model & Data Augmentation & Type & BLEU & AL & AP & DAL \\
    \hline
    \hline
    EN→DE & & & & & & \\
    \hline
    Wait-K \citep{DBLP:conf/ijcnlp/MaPK20} & $\times$ &  End-to-End & 13.95 & 1.75 &  0.79 & 1.98 \\
    CAAT\citep{DBLP:conf/emnlp/LiuDLLC21} & $\times$ &  End-to-End & 22.1 & 1.92 &  - & - \\
    \citet{DBLP:conf/iwslt/WangGLQWLSCZTYQ22} & \checkmark &  End-to-End & 22.13 & 2.37 &  0.86 & 2.52 \\
    \citet{DBLP:conf/iwslt/LiuDLHD21} & \checkmark &  Cascaded & 29.68 & 1.86 &  0.82 & 2.65 \\
    \citet{DBLP:conf/iwslt/PolakPNLMNBW22} & \checkmark &  End-to-End & 31.47 & 1.93 & 0.86 & 2.96 \\
    \hline
    Ours & \checkmark & End-to-End & \textbf{31.69} & \textbf{1.92} & 0.77 & 2.63 \\
      & \checkmark & Cascaded & \textbf{33.54} & \textbf{1.88} & 0.83 & 2.84 \\
    \hline
    \hline
    EN→JA & & & & & & \\
    \hline
    \citet{DBLP:conf/iwslt/WangGLQWLSCZTYQ22} & \checkmark & End-to-End & 12.82 & 1.84	& 0.94 & 3.37 \\
    \citet{DBLP:conf/iwslt/PolakPNLMNBW22} & \checkmark & End-to-End & 16.92 & 2.46	& 0.9 & 3.22 \\
    \hline
    Ours & \checkmark & End-to-End & 16.28 & 1.86 & 0.81 & 2.45 \\
     & \checkmark & Cascaded & \textbf{17.89} & \textbf{1.98} & 0.83 & 2.89 \\
    \hline
    \hline
    EN→ZH & & & & & & \\
    \hline
    \citet{DBLP:conf/iwslt/WangGLQWLSCZTYQ22} & \checkmark & End-to-End & 20.38 & 1.75	& 0.94 & 3.34 \\
    \citet{DBLP:conf/iwslt/PolakPNLMNBW22} & \checkmark & End-to-End & 23.61 & 1.75	& 0.85 & 2.56 \\
    \citet{DBLP:conf/iwslt/ZhuWLZCZMWY22} & \checkmark & End-to-End & 22.49 & 1.27	& - & - \\
     & \checkmark & Cascaded & 25.87 & 1.99	& 0.87 & 3.35 \\
    \hline
    Ours & \checkmark & End-to-End & \textbf{24.36} & \textbf{1.87} & 0.92 & 2.68 \\
     & \checkmark & Cascaded & \textbf{27.23} & \textbf{1.98} & 0.83 & 2.89 \\
    \hline
    \hline
    \end{tabular}
    }
    \caption{Main results comparing with other work. * \citet{DBLP:conf/iwslt/LiuDLHD21} is the winning submission of the EN→DE direction for the IWSLT 2021 simulST track. \citet{DBLP:conf/iwslt/PolakPNLMNBW22} is the winning submission of the EN→DE and EN→JA directions for the IWSLT 2022 simulST track. \citet{DBLP:conf/iwslt/ZhuWLZCZMWY22} is the winning submission of the EN→ZH direction for the IWSLT 2022 simulST track.}
    \label{tab:results}
\end{table*}

\subsection{Pre-trained Models}

\paragraph{Wav2Vec 2.0} Wav2Vec 2.0 \citep{DBLP:conf/nips/BaevskiZMA20} is an advanced Transformer encoder model that can receive raw waveforms as input and generate high-level representations. Its architecture includes two distinct components: first, a convolution-based feature extractor that downsamples long audio waveforms into features with similar lengths to spectrograms; secondly, a deep Transformer encoder that utilizes self-attention and feed-forward neural network blocks for transforming the features without any further downsampling. Throughout the self-supervised training process, the network trains using a contrastive learning strategy, whereby the already downsampled features are randomly masked, and the model learns to predict the quantized latent representation of the masked time step.


\paragraph{mBart50} mBart50 \citep{DBLP:journals/corr/abs-2008-00401} is a Transformer-based language model that operates as an encoder-decoder architecture. During training, the model differs from typical language modeling settings, where it predicts the next word in the sequence. Instead, it is designed to reconstruct a sequence from its noisy version. The model was extended to a multilingual version, where corpora from multiple languages were combined during training. The mBart50 version is pretrained on 50 different languages.


\subsection{Cascade System}

\subsubsection{ASR} We utilized wenet \citep{DBLP:conf/interspeech/ZhangWPSY00YP022} to implement the U2 \citep{DBLP:journals/corr/abs-2012-05481} model, initializing the U2 encoder parameters with Wav2Vec 2.0 \citep{DBLP:conf/nips/BaevskiZMA20}. To tokenize ASR texts, we employed Sentencepiece and developed a vocabulary of up to 20,000 sub-tokens. For training the ASR model, we set the batch size to a maximum of 40,000 frames per card and implemented an inverse square root for $lr$ scheduling with warm-up steps set at 10,000 and peak $lr$ set at $5e-4$. We optimized the model using Adam \citep{DBLP:journals/corr/KingmaB14} and trained it on 8 V100 GPUs while averaging the parameters over the last 4 epochs to enhance accuracy. Additionally, we augmented all audio inputs with spectral augmentation and normalized them by utterance cepstral mean and variance normalization.


\subsubsection{MT}
To achieve multilingual translation, we fine-tuned mBart50 \citep{DBLP:journals/corr/abs-2008-00401} models using Fairseq \citep{DBLP:journals/corr/abs-1904-01038}, an opEN→source tool for training. Our main parameters included training each model using 8 GPUs, with a batch size of 2048 and a parameter update frequency of 2. We set the learning rate at $5e-5$, used a label smoothing value of 0.1, applied a dropout of 0.2, and incorporated 4000 warmup steps. For optimization, we employed the Adam \citep{DBLP:journals/corr/KingmaB14} optimizer with $\beta1$ = 0.9 and $\beta2$ = 0.98. During inference, we utilized a beam size of 8 and set the length penalties to a value of 1.0.


\subsection{End-to-End System}
Our End-to-End ST model is built by combining the mBart50 \citep{DBLP:journals/corr/abs-2008-00401} decoder with the Wav2Vec 2.0 \citep{DBLP:conf/nips/BaevskiZMA20} encoder, resulting in a robust solution. During the fine-tuning phase, the cross-attention layers that connect the decoder with the encoder require extensive tuning due to the modality mismatch between pretraining and fine-tuning. With the exception of MuST-C \citep{DBLP:conf/naacl/GangiCBNT19}, all datasets contain only speech and transcription data. For these datasets, we leverage the best MT model to generate target pseudo sentences, thereby augmenting the existing data. Our End-to-End ST model is first fine-tuned on the augmented data until optimal coverage is achieved. Subsequently, we proceed to fine-tune the model on the MuST-C \citep{DBLP:conf/naacl/GangiCBNT19} dataset for further refinement.


\subsection{Evaluation}
Our models are evaluated using SimulEval \citep{DBLP:conf/emnlp/MaDWGP20}, a reliable evaluation tool. Translation quality is assessed using detokenized case-sensitive BLEU \citep{DBLP:conf/acl/PapineniRWZ02}, while latency is measured using an adapted version of word-level Average Lagging (AL) \citep{DBLP:conf/acl/MaHXZLZZHLLWW19}. As per the latest IWSLT 2023 setup, an AL score below 2 seconds is classified as Low-Latency.


\section{Results}

\subsection{Main Results}

From Table \ref{tab:results}, we can see that the our systems work well on various language pairs. Our best systems beat the best IWSLT 22 systems of all directions.

\subsection{Comparison with Other Policies}

\begin{figure}[!th]
\centering
\includegraphics[width=0.45\textwidth]{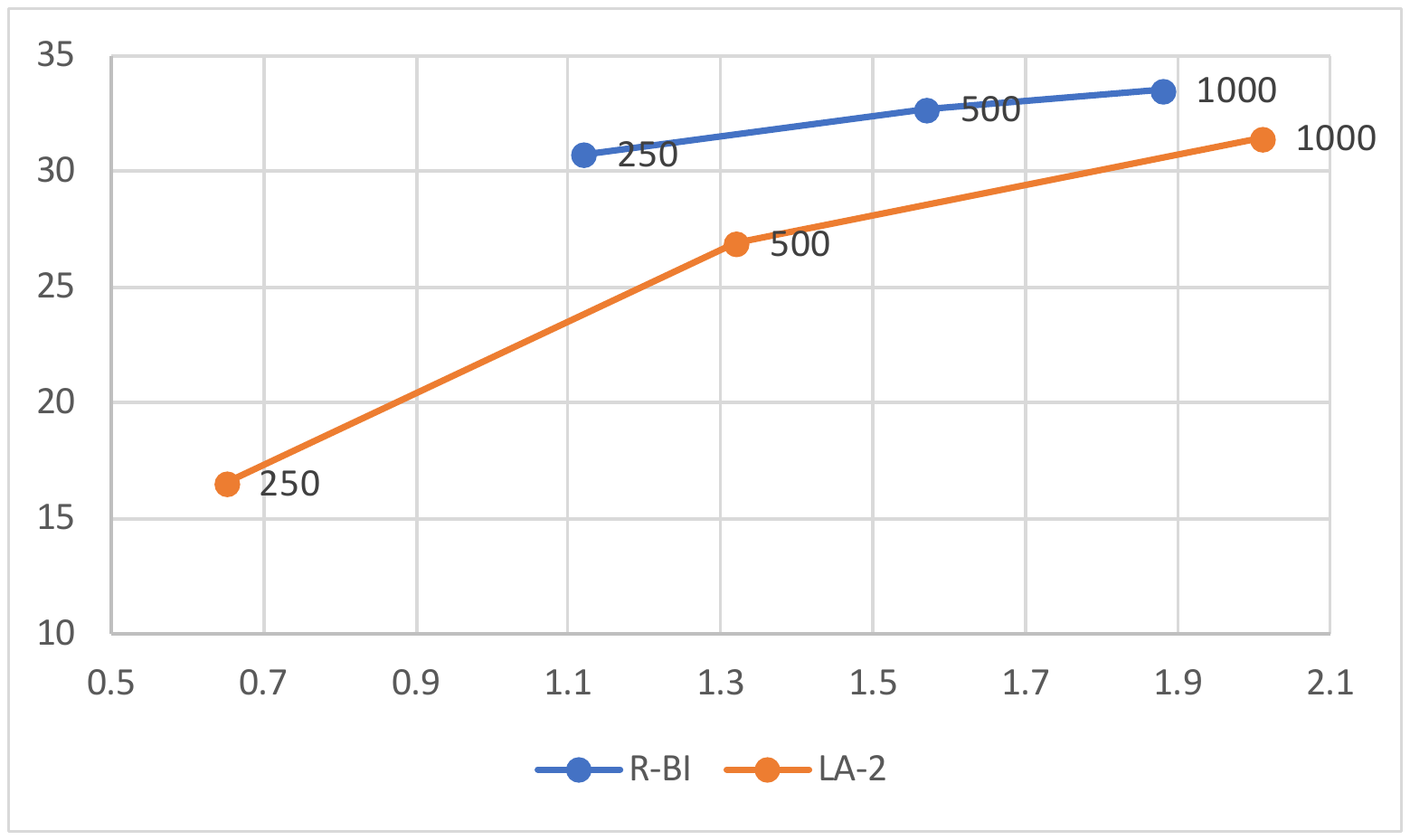} 
\caption{Quality-latency trade-off of different chunk sizes combined with different common prefix decisions policies. The number next to the marks indicates chunk size in milliseconds.}
\label{fig:cp_policy}
\end{figure}

From the Figure \ref{fig:cp_policy}, it can be seen that our R-BI strategy is less sensitive to chunk size compared to LA-2. When the chunk size was reduced from 1000ms to 250ms, our strategy only decreased BLEU by less than 3 points, while LA-2 decreased by as much as 16 points. This indicates that our method is more universally effective.

\subsection{Comparison with OfflineST}

\begin{table}[th]
    \centering
    \begin{tabular}{lccc}
    \hline
    \hline
    Lang & Type & System & BLEU \\
    \hline
    EN→DE & End-to-End & OfflineST & 32.94 \\
    - & - & SimulST & 31.69 \\
    - & Cascaded & OfflineST & 35.23 \\
    - & - & SimulST & 33.54 \\
     \hline
     EN→JA & End-to-End & OfflineST & 18.37 \\
    - & - & SimulST & 16.28 \\
    - & Cascaded & OfflineST & 19.45 \\
    - & - & SimulST & 17.89 \\
     \hline
     EN→ZH & End-to-End & OfflineST & 25.21 \\
    - & - & SimulST & 24.36 \\
    -  & Cascaded & OfflineST & 27.93 \\
    - & - & SimulST & 27.23 \\
    \hline
    \hline
    \end{tabular}
    \caption{Comparison to OfflineST}
    \label{tab:my_label1}
\end{table}


Previous studies have demonstrated that the quality of simultaneous translation can now compete with or even exceed that of offline systems. However, our current research has established a new baseline for the offline system and revealed that there is still a gap of 1-2 BLEU between simultaneous and offline translation. In addition, our findings indicate that cascaded systems continue to outperform end-to-end systems in terms of translation quality, as shown in Table \ref{tab:my_label1}.

\subsection{Ablation Study on different ASR decoding strategies}

\begin{table}[th]
    \centering
    \begin{tabular}{lcc}
    \hline
    \hline
    Lang & Decoding strategies & BLEU \\
    \hline
    EN→DE & ctc\_prefix\_beam\_search & 32.88 \\
    EN→JA & ctc\_prefix\_beam\_search & 16.56 \\
    EN→ZH & ctc\_prefix\_beam\_search & 26.47 \\
    \hline
    EN→DE & attention\_rescoring & 33.54 \\
    EN→JA & attention\_rescoring & 17.89 \\
    EN→ZH & attention\_rescoring & 27.23 \\
    \hline
    \hline
    \end{tabular}
    \caption{Ablation Study of SimulST on different ASR decoding strategies}
    \label{tab:my_label2}
\end{table}

The decoding strategy of "attention\_rescoring" involves using a decoder to re-rank the results based on the decoding output of "ctc\_beam\_search". As a result, "attention\_rescoring" can obtain better ASR results. Table \ref{tab:my_label2} demonstrates that a better ASR decoding strategy can lead to overall better quality results for the system.

\section{Related Work}
\subsection{Simultaneous Models}

At present, SimulST research is divided into two primary directions. Our work follows the first direction, which emphasizes exploring Simultaneous Policies that can directly apply offline models to simultaneous scenarios. These policies enable real-time translation while maintaining high accuracy and fluency.

The second direction involves building Simultaneous Models, such as Monotonic multi-head attention (MMA) \citep{DBLP:conf/iclr/MaPCPG20}, Cross Attention Augmented Transducer (CAAT) \citep{DBLP:conf/emnlp/LiuDLLC21}, and Two single-path based on MMA (Dual Paths) \citep{DBLP:conf/acl/ZhangF22}. These models are specifically designed to generate incomplete translation results based on incomplete audio/text inputs. They use advanced techniques like online decoding and incremental processing to achieve near-real-time performance.

While both directions have their advantages and limitations, our focus on Simultaneous Policies allows us to leverage the strengths of existing offline models, making it easier to integrate with current translation frameworks and reducing the need for extensive retraining. Additionally, our approach shows promising results in terms of both accuracy and latency, making it a viable option for practical use cases.

\section{Conclusion}

In conclusion, this paper revisited the Incremental Decoding framework for Low-Latency Simultaneous Speech Translation and highlighted the potential risk of errors when the system outputs from incomplete input. To address this issue, we proposed a flexible and effective policy, "Regularized Batched Inputs", which includes two regularization methods to improve input diversity and reduce errors. Our experiments on IWSLT SimulST tasks demonstrate that our proposed methods achieve low latency with no more than a 2 BLEU point loss compared to offline systems while achieving several new SOTA results in multiple language directions. We believe that these findings will be valuable for researchers and practitioners working in the field of simultaneous speech translation and will help to advance the development of more accurate and efficient systems.

\section*{Limitations}
While our method proves to be highly effective in achieving high translation quality in ideal simultaneous translation scenarios, it is important to acknowledge its limitations in real-world settings. Firstly, in real simultaneous translation scenarios, it is often necessary to display the output of automatic speech recognition (ASR) alongside the translation. However, our method does not fundamentally improve the accuracy of ASR results since we do not intervene or adjust the ASR during the translation process. Error correction\citep{10096194} is a simple and effective method that can significantly improve the quality of ASR.

Secondly, ideal simultaneous translation scenarios require smooth and continuous translations without abrupt changes. However, in real-world simultaneous settings, some level of discontinuity may be allowed. Specific studies, like the work by Re-translation\citep{DBLP:conf/iwslt/ArivazhaganCMF20}, have focused on low-discontinuity scenarios. Our method may not be fully applicable in these scenarios, requiring further improvements to better adapt to such variations.

Therefore, it is important to note that while our method demonstrates strong applicability and effectiveness in ideal simultaneous translation scenarios, it does have limitations regarding ASR quality and the handling of translation discontinuity in real-world simultaneous translation environments. Further research and refinement are necessary to enhance the robustness and feasibility of our method in various complex scenarios.

\bibliography{anthology,custom}
\bibliographystyle{acl_natbib}

\appendix

\end{document}